\documentclass{article}
\usepackage[nonatbib, final]{neurips_2024}

\usepackage[utf8]{inputenc}
\usepackage[T1]{fontenc}
\DeclareUnicodeCharacter{2212}{-}

\usepackage{microtype}
\usepackage{graphicx}
\usepackage{subfigure}
\usepackage{booktabs} %

\usepackage[pdfa]{hyperref}
\usepackage{amsmath}
\usepackage{amssymb}  
\usepackage{amsthm}
\usepackage[dvipsnames]{xcolor} %

\usepackage[font=small]{caption}

\usepackage{wrapfig}
\usepackage{tabto}
\usepackage{colortbl}
\usepackage{pifont}

\usepackage[
backend=biber,
style=ieee,
citestyle=ieee,
maxbibnames=15
]{biblatex}
\definecolor{almond}{HTML}{E5F1D4}

\theoremstyle{plain}

\theoremstyle{remark}

\definecolor{crosscolor}{rgb}{0.969,0.580,0.114} %
\definecolor{checkcolor}{rgb}{0.485,0.640,0.204} %

\def\cmark{{\color{checkcolor}\ding{52}}}
\def\xmark{{\color{crosscolor}\ding{56}}}

\title{Convolutional Differentiable Logic Gate Networks}

\author{Felix Petersen\\
Stanford University\\
Infty\kern0.05emLabs Research\\
{\tt\small mail@felix-petersen.de}
\And
Hilde Kuehne\\
Tuebingen AI Center\\
MIT-IBM Watson AI Lab\\
\kern-1em{\small\texttt{h.kuehne@uni-tuebingen.de}}\kern-1em\\
\AND
Christian Borgelt\\
University of Salzburg\\
{\small\texttt{christian@borgelt.net}}\\
\And
Julian Welzel\\
Infty\kern0.05emLabs Research\\
{\tt\small welzel@inftylabs.com}
\And
Stefano Ermon\\
Stanford University\\
{\tt\small ermon@cs.stanford.edu}
}

\def\etal{{\it et al.}}

\begin{document}
\maketitle

\begin{abstract}
With the increasing inference cost of machine learning models, there is a growing interest in models with fast and efficient inference. 
Recently, an approach for learning logic gate networks directly via a differentiable relaxation was proposed.
Logic gate networks are faster than conventional neural network approaches because their inference only requires logic gate operators such as NAND, OR, and XOR, which are the underlying building blocks of current hardware and can be efficiently executed.
We build on this idea, extending it by deep logic gate tree convolutions, logical OR pooling, and residual initializations. 
This allows scaling logic gate networks up by over one order of magnitude and utilizing the paradigm of convolution.
On CIFAR-10, we achieve an accuracy of 86.29\% using only 61 million logic gates, which improves over the SOTA while being {29$\times$} smaller.
\end{abstract}

\begin{wrapfigure}[18]{r}{0.55\linewidth}
    \centering
    \vspace{-3.6em}
    ~\kern-.35em\includegraphics[width=1.03\linewidth]{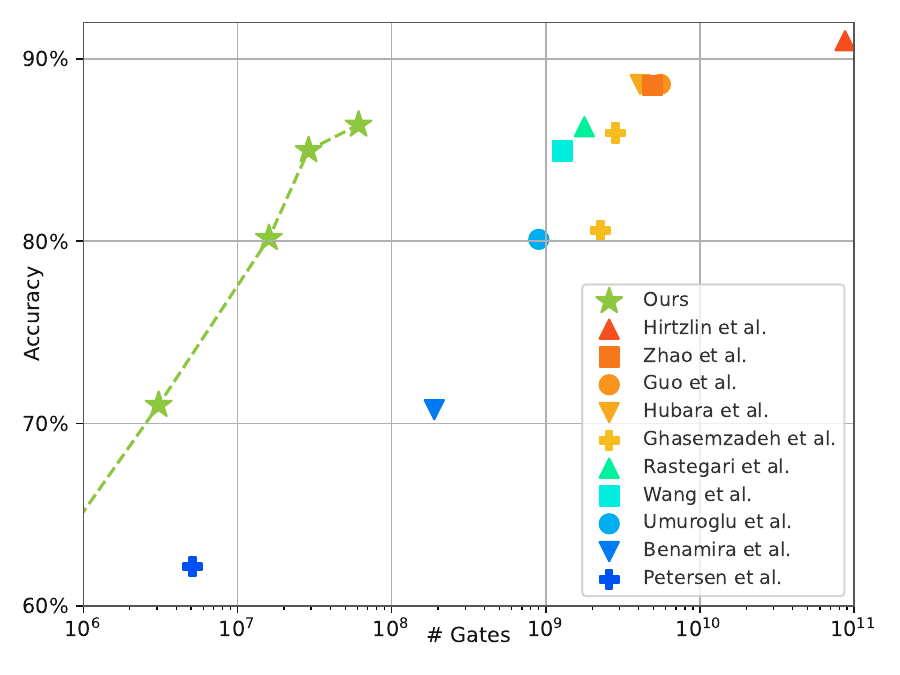}
    \vspace{-1.85em}
    \caption{\label{fig:sota}%
    Gate count vs.~accuracy plot on the CIFAR-10 data set. 
    Our models ({\color{LimeGreen!90!black}$\bigstar$}) are substantially above the pareto-front of the SOTA baselines.
    Gate counts are proportional to chip area.
    Our models are more efficient than the SOTA by {factors of $\geq29\times$.}  %
    Note that the $x$-axis (gate count) is on a log-scale.}
    \vskip-1em
\end{wrapfigure}

\section{Introduction}

Deep learning has led to a variety of new applications, opportunities, and use-cases in machine vision. %
However, this advancement has come with considerable computational and energy costs for inference~\cite{desislavov2021compute}.
Therefore, an array of methods has been developed for efficient deep learning inference~\cite{qin2020binary, gholami2021survey, hoefler2021sparsity, liu2022unreasonable, molchanov2017pruning, petersen2022difflogic}.
These include binary weight neural networks (BNNs)~\cite{qin2020binary}, a set of methods for quantizing neural network weights down to binary representations (and sometimes also binary activations); quantized low-precision neural networks~\cite{gholami2021survey}, a superset of BNNs %
and sparse neural networks~\cite{hoefler2021sparsity, liu2022unreasonable, molchanov2017pruning}, a set of approaches for pruning neural networks and increasing sparsity.
These methods have been successfully utilized for efficient vision model inference.

The state-of-the-art (SOTA) method for small architectures, deep differentiable logic gate networks (LGNs)~\cite{petersen2022difflogic}, approaches efficient machine learning inference from a different direction: learning an LGN (i.e., a network of logic gates such as NAND and XOR) directly via a differentiable relaxation. %

Differentiable LGNs directly learn the combination of logic gates that have to be executed by the hardware. 
This differs from other approaches (like BNNs) that require translating an abstraction (like matrix multiplication-based neural networks) into executable logic for inference, an inductive bias that comes with a considerable computational burden.
By optimizing the logic directly on the lowest possible level instead of optimizing an abstraction, differentiable LGNs lead to very efficient inference on logic gate-based hardware (e.g., CPU, GPU, FPGA, ASIC). 
Recently, differentiable LGNs achieved SOTA inference speeds on MNIST~\cite{lecun2010mnist, petersen2022difflogic}.
However, a crucial limitation was the random choice of connections, preventing LGNs from learning spatial relations, as they arise in images, which limited performance to an accuracy of only $62\%$ on CIFAR-10~\cite{krizhevsky2009cifar10, petersen2022difflogic}.
To address this limitation, we propose to extend differentiable LGNs to convolutions.
Specifically, we propose deep logic gate tree convolutions, i.e., kernels comprised of logic gate trees applied in a convolutional fashion.
Using trees of logic gates, instead of individual gates, increases the expressivity of the architecture while minimizing memory accesses, improving accuracy and accelerating training as well as inference. 
Further, we adapt pooling operations by representing them with logical \textit{or} gates (relaxed via the maximum t-conorm), improving the effectiveness of convolutions in LGNs.
Additionally, we propose ``residual initializations'', a novel initialization scheme for differentiable LGNs that enables scaling them up to deeper networks by providing differentiable residual connections. 
These advances lead to an accuracy of {$86.29\%$} on CIFAR-10 using only {$61$} million logic gates, leading to cost reductions by {$\geq29\times$} compared to SOTAs as displayed in Figure~\ref{fig:sota}.

\section{Background}
\label{sec:background}

\begin{wrapfigure}[17]{r}{.62\linewidth}
    \centering
    \vspace*{-3.7em}
    \includegraphics[width=\linewidth]{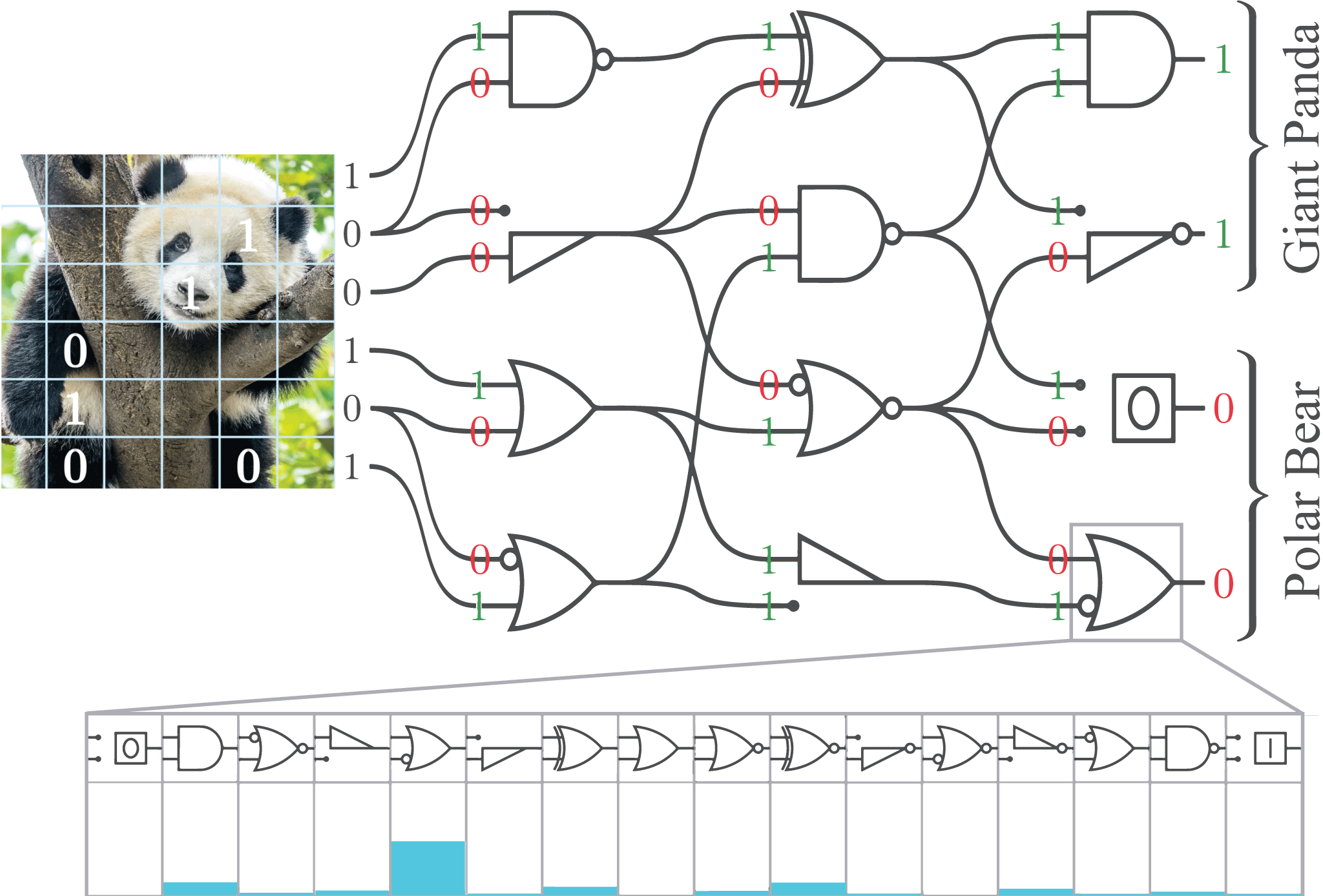}
    \caption{
        Architecture of a randomly connected LGN.
        Each node corresponds to one logic gate. 
        During training, the distribution over choices of logic gates (bottom, 16 options) is learned for each node. 
    }
    \label{fig:rand}
\end{wrapfigure}%
Our work builds on and extends differentiable logic gate networks~\cite{petersen2022difflogic}.
To recap, logic gate networks (LGNs) are networks of nodes that are binary logic gates like AND, NAND, or XOR.
LGNs are also known as binary circuits or logical circuits, and are the format in which any digital hardware is implemented on the lowest pre-transistor abstraction level.
The function that an LGN computes depends on the choices of logic gates that form its nodes and how these nodes are connected.
Optimizing an LGN requires choosing the connections and deciding on a gate for each node.
A primary challenge when optimizing LGNs is that they are, by default, non-differentiable, preventing gradient descent-based training, making this problem conventionally a combinatorial problem. 
However, when applied to machine learning problems, solving the combinatorial problem conventionally becomes infeasible as we require millions of parameters or gates.
Thus, a differentiable relaxation of randomly connected LGNs has been proposed, which allows training LGNs with gradient descent~\cite{petersen2022difflogic}, overcoming the exponential difficulty of optimizing LGNs.
In the remainder of this section, we cover the structure, relaxation, and training of differentiable LGNs, which we also illustrate in Figure~\ref{fig:rand}.

\vspace{-.7em}
\paragraph{Structure} 
LGNs follow a layered structure with each layer comprising a number of nodes, each comprising one logic gate (3 layers with 4 logic gates each in Fig.~\ref{fig:rand}).
As logic gates are inherently non-linear, LGNs do not require any activation functions.
Further, LGNs do not have any weights nor any biases as they do not rely on matrix multiplications.
Due to the binary (i.e., two-input) nature of the nodes, LGNs are necessarily sparse and cannot form fully-connected networks.
The connectivity between nodes has so far been (fully) randomly selected, which works well for easier tasks but can become problematic if there is inherent structure in the data as, e.g., in images.
During training, the connections remain fixed and the learning task comprises the choice of logic gate at each node.

\vspace{-.7em}
\paragraph{Differentiable Relaxation} 
To learn the choices of logic gate for each node with gradient descent requires the network to be differentiable; however, the LGN is by default not differentiable for two reasons:
~(i)~Because a logic gate computes a discrete function of its (Boolean) inputs, it is not differentiable.
~(ii)~Because the choice of logic gate is not a continuous parameter, but a discrete decision, it is not differentiable.
Petersen~\etal~\cite{petersen2022difflogic} propose to differentiably relax each logic gate to real-valued logic via probabilistic logic~\cite{van2020analyzing, klir1997fuzzy}.
For example, a logical \textit{and} ($a_1 \wedge a_2$) is relaxed to $a_1 \cdot a_2$ and a logical \textit{exclusive or} ($a_1 \oplus a_2$) is relaxed to $a_1 + a_2 -2\cdot a_1 \cdot a_2$, which corresponds to the output probability when considering two independent Bernoulli variables with coefficients $a_1, a_2$. 
To make the choice of logic gate learnable, Petersen~\etal~\cite{petersen2022difflogic} introduce a probability distribution over the 16 possible logic gates ($\mathcal{S}$), which is encoded as the softmax of 16 trainable parameters.
For a trainable parameter vector $\mathbf{z}\in\mathbb{R}^{16}$ and all 16 possible logic gate operations as $g_0, ..., g_{15}$, the differentiable logic gate as the expectation over its outputs can be computed in closed-form as 
\begin{equation}
    f_\mathbf{z}(a_1, a_2)
    = \mathbb{E}_{i\sim\mathcal{S}(\mathbf{z})\,,\,A_1\sim\mathcal{B}(a_1)\,,\,A_2\sim\mathcal{B}(a_2)}\Big[ g_i(A_1, A_2) \Big]
    = \sum_{i=0}^{15} \frac{\exp(z_i)}{\sum_j \exp(z_j)}\cdot g_i(a_1, a_2)\,.
    \label{eq:logic-gate-f}
\end{equation}%
With these two ingredients, logic gate networks become end-to-end differentiable.

\vspace{-.7em}
\paragraph{Initialization, Training, and Discretization} 
Training differentiable logic gate networks corresponds to learning the parameters inducing the probability distributions over possible gates.
The parameter vector $\mathbf{z}$ for each node has so far been initialized with a standard Gaussian distribution.
The connections are randomly initialized and remain fixed during training.
For classification tasks, each class is associated with a set of neurons in the output layer and active neurons in each set are counted composing a class score (group sum, right part of Fig.~\ref{fig:rand}).
After dividing them by a temperature $\tau$, the class scores are used as logits in a softmax cross-entropy loss.
Differentiable LGNs perform best when trained with the Adam optimizer~\cite{kingma2015adam}.
Empirical evidence showed that the softmax distributions typically converge to concrete choices of logic gates.
Thus, differentiable LGNs can be discretized to hard LGNs for deployment on hardware by selecting the logic gate with the largest probability.
This discretization process incurs only a minimal loss in accuracy compared to the differentiable LGN~\cite{petersen2022difflogic}.

\vspace{-.7em}
\paragraph{Limitations}
Differentiable LGNs have shown significant limitations wrt.~the available architectural components. 
Previously, they did not provide the option to capture local spatial patterns as they were randomly connected and only operated on flattened inputs~\cite{petersen2022difflogic}.
Further, they previously performed well only up to a depth of $6$ layers~\cite{petersen2022difflogic}. 
Thus, more complex relationships between inputs cannot be modeled.
Finally, while they provide SOTA performance, differentiable LGNs are very computationally expensive to train, e.g., a vanilla 5 million gate network required 90 hours on an A6000 GPU~\cite{petersen2022difflogic}.
In~the following, we address these limitations by introducing convolutional logic tree layers, logical or pooling, residual initializations, as well as computational considerations for scaling.

\section{Convolutional Logic Gate Networks}
\label{sec:convlgn}

\begin{wrapfigure}[19]{r}{0.5\textwidth}
    \centering
    \vspace{-2.9em}
    \includegraphics[width=.9\linewidth]{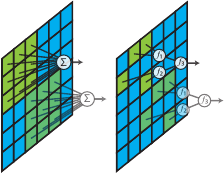}\\[-1.5em]
    \hfill(a)\hfill\hfill~~(b)\hfill~~~~~~
    \caption{\label{fig:convolution}%
        Conventional convolutional neural networks (a) compared to convolutional logic gate networks (b). 
        The images illustrate the first and second to last kernel placements. 
        The nodes correspond to weighted sums (a), and binary logic gates $f_1, f_2, f_3$ (b), respectively.
        The weights / choices of logic gates are shared between kernel placements.
        For visual simplicity, only a single input channel and kernel (output channel) is displayed.
    }
    \vskip-1em
\end{wrapfigure}

Convolutional neural networks (CNNs) have experienced tremendous success, being a core contributor to the current machine learning ascendancy starting with their progress on the ImageNet classification challenge in 2012~\cite{krizhevsky2012imagenet}.
Underlying CNNs is the discrete convolution of an input tensor $\mathbf{A}$ (e.g., an input image or hidden activations) and a linear function / kernel $\mathbf{W}$, denoted as $\mathbf{A} * \mathbf{W}$.
CNNs are especially effective in vision tasks due to the equivariance of the convolution, which allows the network to generalize edge, texture, and shapes in different locations by sharing the parameters at all placements.
However, existing differentiable LGN methods do not support convolutions.
\\[.45em]
In this work, we propose to convolve activations $\mathbf{A}$ with differentiable binary logic gate trees. 
While we could convolve $\mathbf{A}$ with an individual logic gate, we observe that actually convolving $\mathbf{A}$ with a (deep) logic gate network or tree leads to substantially better performance as it allows for greater expressivity of the model. 
Similar to how the inputs to each logic gate are randomly initialized and remain fixed in conventional differentiable LGNs, we randomly construct the connections in our logic gate tree kernel function. 
However, we need to put additional restrictions on the connections for logic gate network kernels.
Specifically, we construct each logic gate network kernel as a complete binary tree of depth $d$ with logic gates as nodes and binary input activations as leaves. 
The output of the logic gate operation is then the input to the next higher node, etc.
To capture spatial patterns, we select the inputs / leaves of the tree from the predefined receptive field of the kernel of size  $s_h\times s_w$. 
Based on the depth of the tree, we randomly select as many inputs as necessary. %
For example, we could construct a binary tree of depth $d=2$, which means that we need to randomly select $2^d=4$ inputs from our receptive field, e.g., of size $64\times3\times3$, which corresponds to $64$ input channels with a kernel size of $3\times3$.
This tree structure allows to capture fixed spatial patterns and correlations beyond pair-wise binary inputs. 
Further, it extends the concept of spacial equivariance to LGNs as such trees can be used as kernel filters, capturing general patterns in different locations.
Using trees of logic gates instead of individual logic gates also has the advantage of reducing memory accesses and improving training and inference efficiency.
We remark that, as we apply convolution, the parameterization of each node is shared between all placements of the kernel (which contrasts convolution from mere local connectivity.)
In Figure~\ref{fig:convolution}, we illustrate the difference between conventional CNN models and convolutional logic gate networks.

During training, the network learns which logic gate operation to choose at each node. 
Thus, each logic tree kernel is parameterized via the choices of each of the $2^d-1$ logic gates, which are learnable.
For a logic kernel of depth~$2$, we call these logic gates $f_1, f_2, f_3$ (or more formally $f_{\mathbf{z}_1}, f_{\mathbf{z}_2}, f_{\mathbf{z}_3}$ for parameter vectors $\mathbf{z}_1, \mathbf{z}_2, \mathbf{z}_3$ corresponding to Equation~\ref{eq:logic-gate-f}).
Given input activations $a_1, a_2, a_3, a_4$, the kernel is expressed as a binary tree of these logic gates:
\begin{equation}
    f_3(\,f_1(a_{1}, a_{2}), f_2(a_{3}, a_{4})\,).
\end{equation}
For an input $\mathbf{A}$ of shape $m\times h\times w$ ($m$ input channels; height; width) and connection index tensors $\mathbf{C}_M, \mathbf{C}_H, \mathbf{C}_W$\footnote{$\mathbf{C}_M \in \{1, ..., m\}^{n\times 4}$ indicates which out of $m$ input channels is selected; $\mathbf{C}_H \in \{1, ..., s_h\}^{n\times 4}$ and $\mathbf{C}_W \in \{1, ..., s_w\}^{n\times 4}$ indicate the selected position inside of the receptive field of size $s_h \times s_w$.}, each of shape $n\times4$ ($n$ tree kernels / channels; $4$ inputs per tree), the output is
\newsavebox{\mybox}
\savebox{\mybox}{$\mathbf{C}_M[k,\! 1],
                  \mathbf{C}_H[k,\! 1]{+}i,
                  \mathbf{C}_W[k,\! 1]{+}j$}
\newdimen\myht \myht=\ht\mybox
\newdimen\mywd \mywd=\wd\mybox
\begin{equation}\label{eq:conv2d}
\begin{aligned}
\!\!\mathbf{A}'[k, i, j] =
& \,\,
      f^{k}_3\big(
      f^{k}_1\big(
      \mathbf{A}\big[\text{\resizebox{.885\mywd}{.95\myht}{$\mathbf{C}_M[k,\! 1],
                     \mathbf{C}_H[k,\! 1]{+}i,
                     \mathbf{C}_W[k,\! 1]{+}j$}}\big],
      \mathbf{A}\big[\text{\resizebox{.885\mywd}{.95\myht}{$\mathbf{C}_M[k,\! 2],
                     \mathbf{C}_H[k,\! 2]{+}i,
                     \mathbf{C}_W[k,\! 2]{+}j$}}\big]
      \big), \\
& \,\,\phantom{f^{k}_3\big(}
      f^{k}_2\big(
      \mathbf{A}\big[\text{\resizebox{.885\mywd}{.95\myht}{$\mathbf{C}_M[k,\! 3],
                     \mathbf{C}_H[k,\! 3]{+}i,
                     \mathbf{C}_W[k,\! 3]{+}j$}}\big],
      \mathbf{A}\big[\text{\resizebox{.885\mywd}{.95\myht}{$\mathbf{C}_M[k,\! 4],
                     \mathbf{C}_H[k,\! 4]{+}i,
                     \mathbf{C}_W[k,\! 4]{+}j$}}\big]
      \big)\big)
\end{aligned}
\end{equation}
for $k \in \{1, ..., n\}$ where $n$ is the number of tree kernels, $i \in \{1, ..., (h-s_h+1)\}$, and $j \in \{1, ..., (w-s_w+1)\}$ where $s_h \times s_w$ is the receptive field size.
Note that, in Equation~\ref{eq:conv2d}, for each output channel~$k$ the logic gates $f^{k}_1, f^{k}_2, f^{k}_3$ (or their relaxed form) are chosen and parameterized independently.
Per convolution, all placements  (indexed via $i, j$) of one kernel share their parameters.

After introducing convolutional LGNs, in the remainder of the section, we introduce our additional components, training strategies, and our architecture.

\subsection{Logical \textit{Or} Pooling}

\begin{wrapfigure}[13]{r}{.489\linewidth}
    \vspace{-3.3em}
    ~\kern-.9em\includegraphics[width=1.08\linewidth]{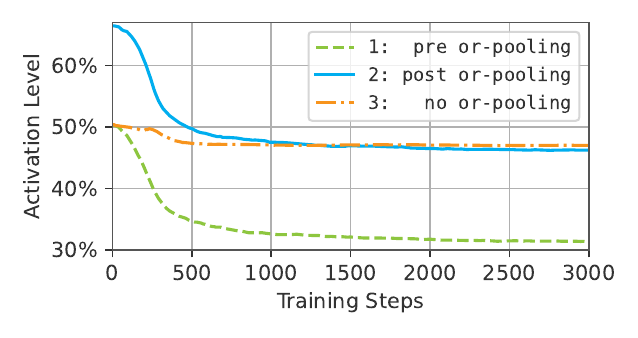}
    \vspace{-2.2em}
    \caption{\label{fig:sparsity}%
        Plot of the density of activations for the second convolutional block of an \textit{or}-pooling based convolutional LGN. 
        It shows that training implicitly enforces that the outputs of the block have the activation level of a no-pooling network (i.e., with pure stride).
    }
\end{wrapfigure}
In CNNs, max-pooling is a crucial component selecting the largest possible activation over a predefined receptive field, e.g., for $2\times2$, $\max(a_{i, j}, a_{i, j+1}, a_{i+1, j}, a_{i+1, j+1})$~\cite{krizhevsky2012imagenet}.
To adopt this for logic, we propose to use the disjunction of the binary activations $a_{i, j} \lor a_{i, j+1} \lor a_{i+1, j} \lor a_{i+1, j+1}$ via the logical \textit{or}.
Instead of using a probabilistic relaxation of the logical \textit{or}, we can use the maximum t-conorm relaxation of the logical \textit{or} ($\bot_{\max}(a,b)=\max(a,b)$). 
By setting the stride of the pooling operation to the size of its receptive field, this has 
a range of crucial computational advantages: (i) it is faster to compute than probabilistic relaxation; (ii) we only need to store the maximum activation and index; (iii)~we only need to backpropagate through the maximum activations during training. 

Intuitively, using many logical $or$s could lead to the outputs of the activations becoming predominantly~$1$.
However, we find that, during training, this is not an issue as using \textit{or} pooling causes an automatic reduction of pre-pooling activations, resolving this potential concern.
This phenomenon is shown in Figure~\ref{fig:sparsity}.
Here, the average activation of a convolutional block of a logic network with $2\times2$ strided \textit{or} pooling is illustrated. 
For a random network without pooling, we expect and observe an average activation of 50\% (dash-dotted).
We observe that the post \textit{or} pooling activations (solid line) for the initialized models is 66.5\%, which follows expectation. 
The pre \textit{or} pooling activations (dashed) are initialized at 50\%, also following expectations.
With training, the post \textit{or} pooling activations (solid) rapidly converge to the average activations of a network without pooling, preventing any problematic saturation of activations.
We do not introduce any explicit regularization enforcing this behavior, but instead found this to be an emerging behavior of training. %

\subsection{Residual Initialization}
The parameters $\mathbf{z}$ of existing differentiable LGNs were initialized as random draws from a Gaussian distribution.
Unfortunately, after applying softmax, this leads to rather ``washed out'' probability distributions over choices of logic gates.
Accordingly, the expected activations, as computed via Equation~\ref{eq:logic-gate-f}, are also washed out, quickly converging towards $0.5$ in deeper networks.
This also leads to vanishing gradients in existing differentiable LGNs:
With Gaussian initialization, during backpropagation, the gradient norm decays at each logic gate by a factor between $0.1$ and $0.2$ for an initialized network, exponentially slowing training for deeper networks.

In CNNs, a technique for preventing vanishing gradients and preventing loss of information in deep networks are residual connections. 
Residual connections conventionally add the input to a block to the output of this block~\cite{he2016deep}.
However, when operating in logic, we cannot perform such additions.

\begin{wrapfigure}[8]{r}{.725\linewidth}
    \centering
    \vspace{-2.1em}
    \includegraphics[width=\linewidth]{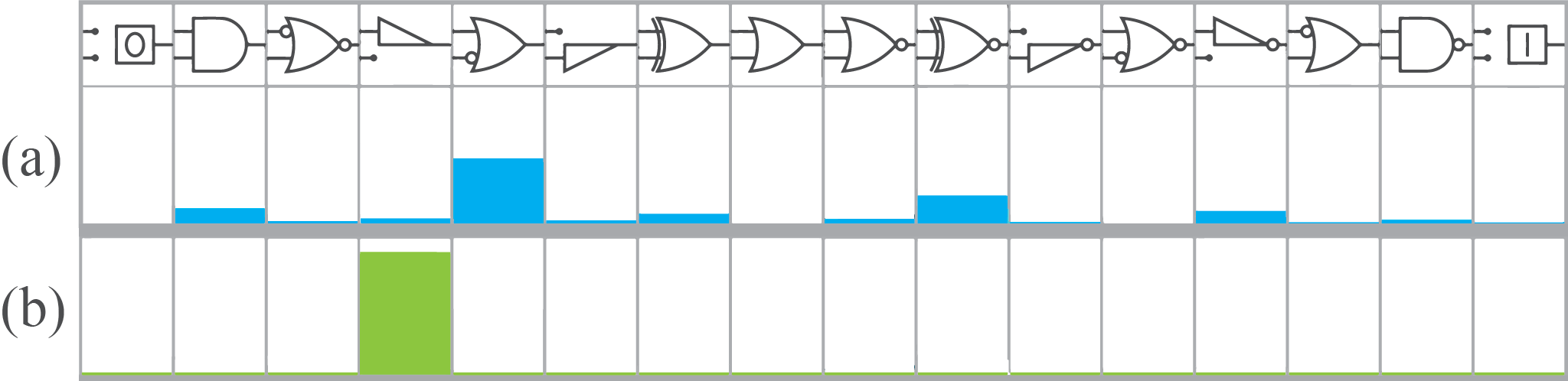}
    \caption{
        Gaussian initialization (a)~\cite{petersen2022difflogic} vs.\ our residual initialization (b).\label{fig:residual-init-concept}
    }
\end{wrapfigure}

To prevent the loss of information through washed out activations and reduce vanishing gradients with a joint strategy, we propose \textit{residual initializations}.
For this, we initialize each logic gate not randomly but instead to be primarily a feedforwarding logic gate.
Here, we choose `$A$' as a canonical choice and choosing `$B$' would be equivalent.
In our experiments, we found that initializing the probability for the logic gate choice `$A$' to around $90\%$ and setting all other gates to $0.67\%$ works well.
This corresponds to setting the parameter $z_3 = 5$ and all other $z_i=0$ for $i\neq 3$ in accordance to Eq.~\ref{eq:logic-gate-f}.
We illustrate an example of residual initializations compared to the existing Gaussian initializations in Figure~\ref{fig:residual-init-concept}.

Residual initializations prevent the loss of information as well as vanishing gradients in deeper networks.
During training, whenever a residual connection is not required, the model learns to replace the feedforward logic gate choice by an actual operation.
Thus, residual initializations are effectively a differentiable form of residual connections that does not require any hard-wiring.
This also means that this form of residuals does not require additional logic gates for residuals.
Residual initializations enable, for the first time, efficient and effective training of LGNs beyond 6 layers.

\subsection{Computational Training Considerations}

Using trees and pooling allows for substantially improved computational training efficiency and memory requirement reductions.
This is because it allows intermediate activations to be used only by the current logic gate tree and because we only need to backpropagate through the maximum activations during \textit{or} pooling.
For example, using learnable trees with a depth of $2$ and \textit{or} pooling with a kernel size and stride of $2\times2$ corresponds to a logic gate tree of depth $2+2=4$ (2 levels are learnable + 2 from pooling) with $16$ inputs and only a single output.
For training, it is most efficient to discard all intermediate values and only store the output and information of which path through the pooling was selected, and during backward to recompute only this path, thereby reducing memory accesses.
The reason for this is that training speed is limited by memory bandwidth and scalability is limited by GPU memory.
On average, this strategy reduces memory accesses by 68\% and reduces the memory footprint by 90\% during training.
For using LGNs in hardware designs, trees and pooling improve the locality of operations and routing, which also leads to more efficient chip layouts.

The residual initializations provide a bias towards the feedforward logic gate in trained LGNs.
As feedforward gates only require a wire and no transistors, this further reduces the necessary transistor count for hardware implementations of the LGNs, reducing the required chip area.

We developed efficient fully-fused low-level CUDA kernels, which, for the first time, enable training of convolutional LGNs. 
The speed of our convolutional layer is up to 200$\times$ faster per logic gate than existing randomly connected LGN implementations~\cite{petersen2022difflogic}.
We will make the code publicly available by including it into the \texttt{difflogic} library at \texttt{\color{blue!60!black}\href{https://github.com/Felix-Petersen/difflogic}{github.com/Felix-Petersen/difflogic}}.

\subsection{LogicTreeNet Architecture}

\begin{wrapfigure}[37]{r}{.49\textwidth}
    \centering
    \vspace{-2.2em}
    \includegraphics[width=1.075\linewidth]{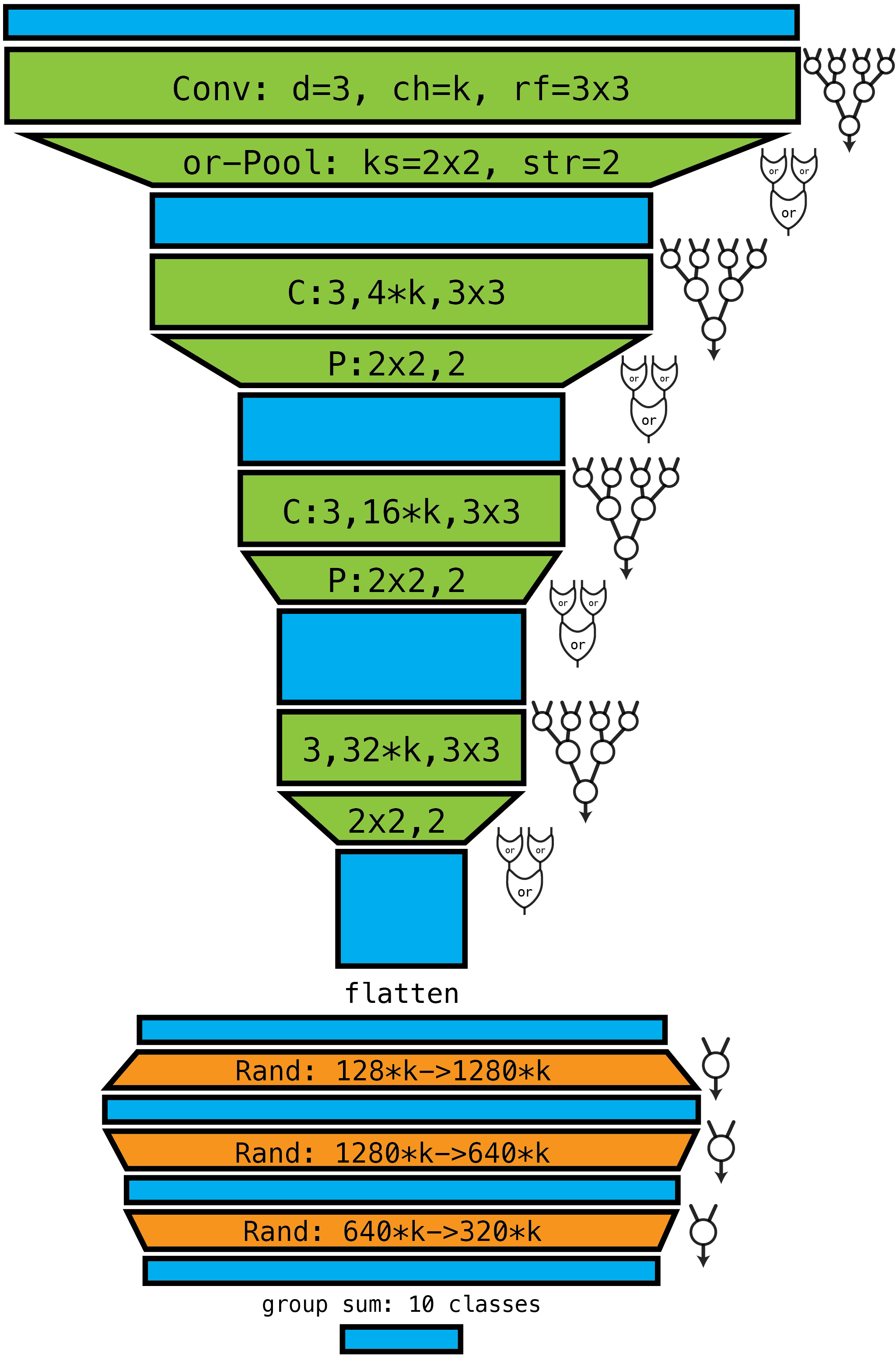}
    \caption{\label{fig:arch}%
        LogicTreeNet architecture.
        The logical architectures of the layers / blocks are illustrated on a per neuron basis.
        Circles indicate a logic gate that can be learned while the logical \textit{or}s remain fixed.
        During training, for the trainable nodes, we use probabilistic relaxations of logic gates, which we parameterize via a softmax distribution over operators (Eq.~\ref{eq:logic-gate-f}/\ref{eq:conv2d}). For the fixed logical \textit{or}s, we use the continuous maximum t-conorm relaxation.
    }
    \vskip-1em
\end{wrapfigure}
In the following, we discuss the design of our convolutional logic gate tree network architectures (LogicTreeNet) for CIFAR-10, which we illustrate in Figure~\ref{fig:arch}. 
We follow the pattern of conventional convolutional architectures and design the architecture by applying convolutional blocks with pooling at the end of each block.
Each block reduces the size by a factor of $2\times2$ and we apply blocks until we reach a size of $2\times2$, increasing the number of channels in each stage.
Following this, we apply two randomly connected layers and a group sum as our classification head.
This architecture has an overall logical depth of $23$ layers, including 4 convolutional blocks (\texttt{Conv}) with tree depths of $d=3$, 4 \textit{or} pooling layers (\texttt{or-Pool}), and 3 randomly connected layers (\texttt{Rand}). 
$15$ of these layers are trainable (\texttt{Conv} blocks and \texttt{Rand}), and the pooling layers remain fixed.
The architecture is defined in terms of a hyperparameter \texttt{k}, which controls the width of the overall network; we consider $\texttt{k}\in\{\text{S}\to32,$ $\text{M}\to256,$ $\text{B}\to512,$ $\text{L}\to1\,024,$ $\text{G}\to2\,048\}$.
In Appendix~\ref{apx:model}, we describe LogicTreeNet layer-by-layer and include a LogicTreeNet for MNIST.

An additional architecture choice is the connectivity for the inputs to a convolutional tree.
While we rely on random choices for the inputs, we restrict the choices of channels ($\mathbf{C}_M$) such that each tree observes only 2 (rather than up to 8) input channels.
This has the two advantages of enforcing spatial comparisons of values within one channel and is more efficient in hardware circuit designs.
When creating hardware designs, for larger models, routing could become a problem due to congestion when connections between channels follow an arbitrary order.
Thus, we restrict the connections between channels to ensure proper routing: we split the model into $\texttt{k}/8$ groups, ensuring no cross-connections between the channels of each group.
This restriction as well as similar hardware specific routing restrictions can be implemented without affecting the accuracy due to the sparsity of the logic gate network model.

\subsection{Input Processing}
For our smaller CIFAR-10 models (S, M), we use 2 bit precision inputs, and encode them using 3 thresholds as in~\cite{petersen2022difflogic}. 
For our larger CIFAR-10 models (B, L, G), we use 5 bit precision inputs, and process them with low-level feature detectors, in particular, we use edge and curvature detector kernels with thresholds, converting them into binary encodings, which are converted into LGNs and not learned.
We note that the gates for the input preprocessing are included in each of the gate counts.

\clearpage

\section{Related Work}
\label{sec:rel-wor}

Beyond differentiable LGNs~\cite{petersen2022difflogic, petersen2022wipo} (covered in Section~\ref{sec:background}), the related work comprises truth table networks~\cite{chatterjee2018learning, benamira2023scalable}, binary and quantized neural networks~\cite{qin2020binary, gholami2021survey}, and sparse neural networks~\cite{hoefler2021sparsity}.

\vspace{-.5em}
\paragraph{Lookup / Truth Table Networks}
Lookup table networks (aka.~truth table networks) are networks comprised of lookup tables (LUTs) or equivalently (potentially complex) logic gates with $n$ inputs.

There are different approaches for learning or constructing lookup table networks.
Chatterjee~\cite{chatterjee2018learning} constructs truth table networks by ``memorizing'' training data in an explorative work to consider relations between memorization and generalization.
Wang~\etal~\cite{wang2019lutnet_fccm, wang2020lutnet_tc} replace the multiplication in BNNs by lookup tables (LUTNet).
Benamira~\etal~\cite{benamira2023scalable} transform Heaviside step function activated CNNs into lookup tables by expressing the binary activation of each neuron via a lookup table that implicitly encodes the weight matrix (TTNet).
This allows obtaining the binary activation of a neuron by ``looking up'' a value from the truth table at a location encoded via the binary inputs of the layer.
Benamira~\etal~\cite{benamira2023scalable} use this as an intermediate representation to then convert the truth tables into LGNs via CNF/DNF (conjunctive / disjunctive normal form) conversion.
The resulting LGNs allow for efficient and effective formal verification.
These resulting LGNs differ from the LGNs considered in this work because they are derived from a conventional CNN and not directly learned, thereby having the inductive bias of the neural network architecture (matmul) and its computational overhead, which is similar to BNNs converted into LGNs.
We remark that, while TTNets are LGNs, TTNets are not differentiable LGNs as there is no differentiable representation of LGNs involved.
Recently, Bacellar~\etal~\cite{bacellar2024differentiable} extended differentiable LGNs to learning logic gates with more than two inputs.

\vspace{-.5em}
\paragraph{Binary and Quantized Low-Precision Networks}

BNNs and quantized neural networks reduce the precision of the weight matrices of a neural network.
For example, BNNs typically use the weights $-1$ and $+1$, but variations are possible.
For quantized neural networks, a popular choice is 8-bit and other options (such as 4-bit~\cite{choi2018pact}) are covered in the literature.
This leads to substantially reduced storage requirements of neural networks at the cost of some accuracy.
Instead of naïvely quantizing weights, these approaches typically involve, e.g., quantization-aware fine-tuning~\cite{gholami2021survey}.
In addition, for some methods, BNNs and quantized neural networks also reduce the precision of the computations and activations, leading to speedups during inference~\cite{qin2020binary, gholami2021survey}.
These approaches typically start with a conventional pre-trained neural network and then convert it into a low-precision representation.
BNNs are among the fastest approaches for efficient inference~\cite{qin2020binary}.

While BNNs (with binary activations, e.g., XNOR-Net~\cite{rastegari2016xnor}) are converted into LGNs for inference on hardware (e.g., on FPGAs~\cite{umuroglu2017finn}), the resulting architectures are fundamentally different from directly trained logic gate networks. 
BNNs have weight matrices and require multiply-accumulate (MAC) operations to express matrix multiplications.
Asymptotically, each MAC requires 8 logic gates while at the same time (with only 2 possible states of the weight) this leads to a smaller expressivity compared to a single learned logic gate (with 16 possible states).
We include a technical discussion in the appendix.
While it is disadvantageous for inference, for training, BNNs have the advantage of operating on a higher abstraction level, simplifying training and allowing for translation between conventional neural networks and~BNNs.
We remark that BNNs with binary input activations and binary weight quantization frequently do not use binary output activations~\cite{santos2023redbit}, which means that only the multiplications within a matrix multiplication are binary, while the remainder of the respective architectures can require floating precision.
In contrast to BNNs, differentiable LGNs are not parameterized via weight matrices but instead via the choices of logic gates at each node~\cite{petersen2022difflogic}.

\vspace{-.5em}
\paragraph{Sparse Neural Networks}

Sparse neural networks are networks that are not densely connected but instead have only selected connections between layers~\cite{hoefler2021sparsity, liu2023don, liu2023more}.
Conceptually, this means multiplying a weight matrix with a binary mask, setting a selection of weights to $0$. 
Sparse nets can be utilized for efficient inference as the sparsity greatly reduces the number of floating-point operations that have to be executed.
For an overview of sparse neural networks, we refer to Hoefler~\etal~\cite{hoefler2021sparsity}.

Due to the binary (i.e., two-input) nature of logic gates, logic gate networks are intrinsically sparse.
Thus, LGNs can be seen as sparse networks; however, sparse neural networks are typically not LGNs and typically operate on real values instead of Boolean values.
As differentiable LGNs use randomly initialized and fixed connections, it is perhaps important to mention that choosing randomly initialized and fixed connections has been shown to also work well for conventional sparse neural networks~\cite{liu2022unreasonable}.

\section{Experiments}

\subsection{CIFAR-10}

\begin{wraptable}[25]{r}{.51\linewidth}
    \centering\setlength{\tabcolsep}{4.5pt}
    \vspace{-1.3em}
    \caption{\label{tab:cifar}%
        \textbf{Main results} for the CIFAR-10 experiments. 
        Our LogicTreeNet models reduce the required numbers of logic gates by factors of {$\geq29\times$} compared to the state-of-the-art models.\ Our\,models\,are\,scaled\,to\,match\,accuracies.
    } %
    \scalebox{.95}{
    \begin{tabular}{lrr}
    \toprule
        Method                                                  & Acc.              & \# Gates \\
    \midrule
        DiffLogic Net (medium)  \cite{petersen2022difflogic}  & {57.39}\%      & 0.51 M    \\  %
        DiffLogic Net (largest) \cite{petersen2022difflogic}  & 62.14\%        & 5.12 M    \\  %
    \midrule
        Conv.~TTNet (small)~\cite{benamira2023scalable}         & 50.10\%        & 0.57 M      \\
        Conv.~TTNet (large)~\cite{benamira2023scalable}         & \textbf{70.75}\%        & 189 M  \\
        FINN CNV~\cite{umuroglu2017finn}                        & \textbf{80.10}\%        & 901 M     \\
        LUTNet~\cite{wang2020lutnet_tc}                         & \textbf{84.95}\%        & 1\,290 M  \\
        XNOR-Net~\cite{rastegari2016xnor} (NIN)~\cite{yu2023xnor} & \textbf{86.28}\%      & 1\,780 M  \\
        RebNet~(1 residual)~\cite{ghasemzadeh2018rebnet}        & 80.59\%        & 2\,270 M  \\
        RebNet~(2 residuals)~\cite{ghasemzadeh2018rebnet}       & 85.94\%        & 2\,830 M  \\
        BinaryNet~\cite{hubara2016binarized}                    & 88.60\%        & 4\,090 M  \\
        Zhao~et al.~\cite{zhao-bnn-fpga2017}                    & 88.54\%        & 4\,940 M  \\
        FBNA CNV~\cite{guo2018fbna}                             & 88.61\%        & 5\,540 M  \\
        Hirtzlin~et al.~\cite{hirtzlin2019stochastic}           & 91.~~~~\%      &87\,400 M  \\
    \midrule
\rowcolor{almond} LogicTreeNet-S         & \textbf{60.38}\%           & 0.40 M \\  %
\rowcolor{almond} LogicTreeNet-M         & \textbf{71.01}\%           & 3.08 M \\  %
\rowcolor{almond} LogicTreeNet-B         & \textbf{80.17}\%           & 16.0 M \\ %
\rowcolor{almond} LogicTreeNet-L         & \textbf{84.99}\%           & 28.9 M \\
\rowcolor{almond} LogicTreeNet-G         & {\textbf{86.29}}\%     & 61.0 M \\  %
    \bottomrule
    \end{tabular}
    }
\end{wraptable}

We train five sizes of LogicTreeNets on the CIFAR-10 data set~\cite{krizhevsky2009cifar10} using the AdamW optimizer~\cite{kingma2015adam, loshchilov2019decoupled} with a batch size of $128$ at a learning rate of $0.02$.
Additional training details and hyperparameters are in Appendix~\ref{apx:training}.
We report our main results in Table~\ref{tab:cifar} and Figure~\ref{fig:sota}.
Our primary evaluation is with respect to the number of logic gates (bin.~ops), which corresponds to the cost in hardware implementations and is proportional to transistor count chip area for ASICs or occupancy on FPGAs.

Comparing our model~(M) with 3.08~M gates to the large TTNet model~\cite{benamira2023scalable}, we can observe that, while the accuracies are similar, our model requires only 1.6\% of the number of logic gates.
Increasing the model size, our model~(B) matches the accuracy of FINN~\cite{umuroglu2017finn}, while requiring only 16~M gates compared to 901 M gates, a $56\times$ reduction.
Considering an even larger variant of our model (L) with 28.9 M~gates, we achieve 84.99\%. 
The smallest baseline model that achieves comparable accuracy (84.95\%) is LUTNet~\cite{wang2020lutnet_tc}, which requires $44.6\times$ as many logic gates.
Finally, considering our largest model (G) with 61~M logic gates, we achieve {$86.29\%$} test accuracy.
We match the accuracy of the Network-in-Network~\cite{yu2023xnor} XNOR-Net~\cite{rastegari2016xnor}, while this baseline requires {$29\times$} as many gates.
Indeed, all networks in the literature below 4 billion gates perform worse than our {61} million gate network.

\begin{wraptable}[17]{r}{.47\linewidth}
    \centering\setlength{\tabcolsep}{4.5pt}
    \centering
    \vspace{-1.25em}
    \caption{\label{tab:cifar-times-fpga}%
        Timing results for CIFAR-10.
        The time is per image on an FPGA. 
        We use a Xilinx VU13P FPGA. Our times are bottleneck by the data transfer onto the FPGA.
        `A' indicates the use of an ASIC.
    }
    \scalebox{.95}{%
    \begin{tabular}{lrrr}
    \toprule
        Method                                                  & Acc.          & FPGA t.   \\
    \midrule
        FINN CNV~\cite{umuroglu2017finn}                        & 80.10\%   & 45.6 $\mu$s  \\
        RebNet~(1 residual)~\cite{ghasemzadeh2018rebnet}        & 80.59\%   & 167 $\mu$s   \\
        RebNet~(2 residuals)~\cite{ghasemzadeh2018rebnet}       & 85.94\%   & 333 $\mu$s   \\
        Zhao~et al.~\cite{zhao-bnn-fpga2017}                    & 88.54\%   & 5.94 ms      \\ 
        FBNA CNV~\cite{guo2018fbna}                             & 88.61\%   & 1.92 ms      \\
        FracBNN~\cite{zhang2021fracbnn}                         & 89.10\%   & 356 $\mu$s    \\
        TrueNorth~\cite{esser2016convolutional}                 & 83.41\%   & \kern-5emA: 801$\mu$s  \\ 
    \midrule
\rowcolor{almond} LogicTreeNet-S  & 60.38\% & 9 ns \\
\rowcolor{almond} LogicTreeNet-M  & 71.01\% & 9 ns \\
\rowcolor{almond} LogicTreeNet-B  & 80.17\% & 24 ns \\
    \bottomrule
    \end{tabular}
    }
\end{wraptable}

After covering the performance of the trained models, we demonstrate their applicability in hardware designs on a Xilinx FPGA as a proof-of-concept.
On CIFAR-10 we limit the hardware development up to the base model (B) due to labor cost.
In Table~\ref{tab:cifar-times-fpga}, we report the results.
We can observe a very favorable FPGA timing trade-off compared to previous works.
Indeed, using our model (B) we achieve 80.17\% accuracy, matching the accuracy of the FINN accelerator, but decreasing inference time from 45.6~$\mu$s to {24 ns}.
In other words, our model achieves {41.6} million FPS, whereas the previously fastest FPGA model achieved 22 thousand FPS (even among all models with $\geq$70\%).
Herein, the limitation preventing us from reaching around 500 million FPS is the transfer speed onto the FPGA.
Here, the difference between the smaller models (S \& M) and the larger model (B) is that (S \& M) receive the input at 2 bit precision whereas (B) receives the input at 5 bit precision. 
We want to remark that substantially accelerated speeds or reduced power consumption could be achieved by manufacturing custom hardware such as ASICs; however, this lies out of the scope of this work and is an interesting future research direction.

We remark that all accuracies reported in the main paper are from discretized LGNs, and all gate counts maintain the full convolutional character (no location-based simplifications, e.g., at zero-padding). 
In Appendix~\ref{apx:train-test-discrete}, we include a plot comparing the differentiable training mode accuracy to the discretized inference mode accuracy. 
Further, we refer to Figure~\ref{fig:sota} for a comparison of LogicTreeNet compared to the pareto-front of the state-of-the-art.

\subsection{MNIST}

\begin{wraptable}[19]{r}{0.56\linewidth}
    \centering
    \vspace{-1.5em}
    \setlength{\tabcolsep}{3.5pt}
    \caption{\label{tab:mnist-fpga}%
        Results of the MNIST experiment.
        We use a Xilinx XC7Z045 FPGA, the same device as FINN CNV. 
        All other baselines utilize equivalent or more powerful FPGAs. 
        }
    \scalebox{.95}{%
    \begin{tabular}{lrrr}
    \toprule
        Method                                              & Acc.          &\# Gates      &FPGA t.     \\
    \midrule
        DiffLogic Net (small) \cite{petersen2022difflogic}  & 97.69\%       & 48 K   & ---  \\  
        DiffLogic Net (largest) \cite{petersen2022difflogic}& 98.47\%       & 384 K  & ---    \\
        DWN~\cite{bacellar2024differentiable}               & 98.77\%       & ---    & 45 ns \\
        \midrule
        TTNet (small) \cite{benamira2023scalable}           & 97.23\%       & 46 K   & ---     \\
        TTNet \cite{benamira2023scalable}                   & 98.02\%       & 360 K  & ---     \\
        LUTNet~\cite{wang2020lutnet_tc}                     & 98.01\%       & ---    & 5 ns              \\
        FINN CNV~\cite{umuroglu2017finn}                        & 98.40\%       & 5.28 M & 641 ns            \\
        FINN FCN~\cite{umuroglu2017finn}                        & 98.86\%       & 258 M & ---                \\
        LowBitNN~\cite{zhan2020field}                       & 99.2~~\%        & ---    & 152 $\mu$s        \\  %
        FPGA-NHAP~\cite{liu2022fpga}                        & 97.81\%       & ---    & 4.9 ms            \\  %
        \midrule
\rowcolor{almond}        LogicTreeNet-S    & 98.46\% & 147 K  & 4 ns  \\ %
\rowcolor{almond}        LogicTreeNet-M    & 99.23\% & 566 K  & 5 ns  \\ %
\rowcolor{almond}        LogicTreeNet-L    & 99.35\% & 1.27 M & ---   \\ 
    \bottomrule
    \end{tabular}
    }
\end{wraptable}

We continue our evaluation on MNIST~\cite{lecun2010mnist}.
Here, we use a slightly smaller model architecture with only 3 (instead of 4) convolutional blocks due to the input size of $28\times28$. 
Each convolutional block has a depth of $3$ and, to maintain valid shapes, we use no padding in the first convolutional block. Each block increases the number of channels by a factor of 3.
This network architecture is described in greater detail in Appendix~\ref{apx:model-mnist}.

We display the results for MNIST in Table~\ref{tab:mnist-fpga}.
Here, our models achieves a range of new SOTAs: 
compared to FINN~\cite{umuroglu2017finn}, we can observe that our small model already improves the accuracy while simultaneously decreasing the model size by a factor of $36\times$, and reducing inference time by a factor of $160$.
Our medium model, with $99.23\%$ test accuracy improves over all BNNs in the literature.
When comparing to LowBitNN~\cite{zhan2020field}, a non-binary model, our medium model reduces the inference time by a factor of $30\,000\times$ while still improving accuracy, increasing throughput from $6\,600$ FPS to $200\,000$ FPS.

Within the, ``one-classification-per-cycle'' regime, comparing to LUTNet~\cite{wang2020lutnet_tc}, we decrease the error from $1.99\%$ to $0.77\%$, and we note that the larger FPGA that LUTNet uses should enable placing LogicTreeNet-L ($0.65\%$ error) multiple times, enabling multiple classifications per cycle.

Concluding, our MNIST models are both the most efficient models in the $\geq98\%$ regime and at the same time also the highest accuracy models with an accuracy of up to $99.35\%$.

\begin{wraptable}[7]{r}{0.27\linewidth}
    \centering
    \vspace{-1.25em}
    \setlength{\tabcolsep}{4pt}
    \caption{
    Variances between individual models on MNIST.\label{tab:seed-var}
    }
    \scalebox{.95}{
    \begin{tabular}{lr}
    \toprule
        Model\kern-.5em & Individual accs. \\
    \midrule
        S & $98.21\%\pm0.31\%$ \\
        M & $99.13\%\pm0.11\%$ \\
        L & $99.29\%\pm0.06\%$ \\
    \bottomrule
    \end{tabular}
    }\vspace{.25em}
\end{wraptable}
\paragraph{Variances}
For small models like the small (S) model for MNIST, which has only $16$ kernels in the first layer, variance due to the fixed connectivity can become a significant concern.
Thus, for the small models we train multiple models simultaneously, and use a validation set of 10\,000 images that we hold-out from the training set (not the test set), and based on which we select the final models.
We present the variations before this selection between individual model in Table~\ref{tab:seed-var}. 
We can see that with increasing model size, the variance decreases.

\subsection{Ablation Study}

\begin{wraptable}[16]{r}{0.64\textwidth}%
    \centering\setlength{\tabcolsep}{3.5pt}
    \vspace{-1.25em}
    \caption{\label{tab:ablation}
        Ablation study on CIFAR-10 wrt.\ architectural choices.
    }%
    \scalebox{.95}{%
    \begin{tabular}{lr|rr|cccc}
    \toprule
        Method                                & \kern-4emAccuracy~~  & \rotatebox{90}{\# train.}~\rotatebox{90}{~~~layers} & \rotatebox{90}{\# total}~\rotatebox{90}{~~~layers} & ~~\rotatebox{90}{\textit{or}-pool}~~ & \rotatebox{90}{residual\phantom{g}}~\rotatebox{90}{init.}  & \rotatebox{90}{weight}~\rotatebox{90}{decay} & \rotatebox{90}{2 input}~\rotatebox{90}{channels}    \\
    \midrule
\rowcolor{almond}  LogicTreeNet-L         & 84.99\%       & 15   & 23 & \cmark & \cmark & \cmark & \cmark \\
\midrule
Conv.\ $d$: 1,1,1,1                        & 80.98\%      & 7   &  15 & \cmark & \cmark & \cmark & \cmark  \\
Conv.\ $d$: 1,1,2,2                        & 82.68\%      & 9   &  17 & \cmark & \cmark & \cmark & \cmark  \\
Conv.\ $d$: 2,2,2,2                        & 83.32\%      & 11  &  19 & \cmark & \cmark & \cmark & \cmark  \\
Conv.\ $d$: 2,2,3,3                        & 84.13\%      & 13  &  21 & \cmark & \cmark & \cmark & \cmark  \\
    \midrule
No \textit{or} pooling                    & 81.45\%      & 15   & 15 & \xmark & \cmark & \cmark & \cmark  \\
Gaussian init.                            & 76.18\%      & 15   & 23 & \cmark & \xmark & \cmark & \cmark  \\
No weight decay                           & 83.94\%      & 15   & 23 & \cmark & \cmark & \xmark & \cmark  \\
8 input channels                          & 83.53\%      & 15   & 23 & \cmark & \cmark & \cmark & \xmark  \\
    \bottomrule
    \end{tabular}
    }
\end{wraptable}
To demonstrate the importance of the provided architectural choices, we provide an ablation study in Table~\ref{tab:ablation}.
Here, we observe that using trees, residual initializations, as well as \textit{or} pooling are integral to the performance of convolutional LGNs. We also provide an ablation wrt.\ model depth.

Starting with the model depth ablation, in Table~\ref{tab:ablation}, we can observe that the performance improves with increasing model depth. 
We observe that decreasing the model depth is detrimental to performance. 
We note that shallower models do not directly correspond to reductions in gate counts because, for deeper models, the rates of trivial gate choices like `$A$` that are removed during logic synthesis is significantly higher.

\begin{wrapfigure}[12]{r}{.52\linewidth}
    \centering
    \vspace{-1.75em}
    ~\kern-1em\includegraphics[width=1.06\linewidth]{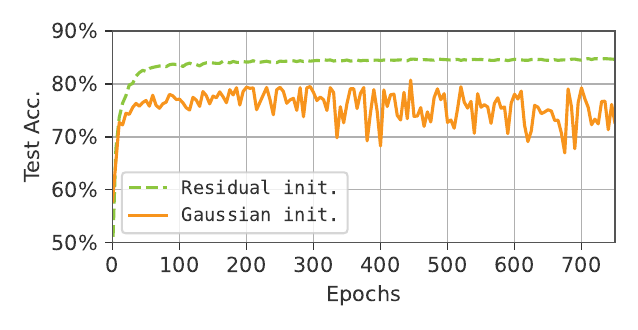}
    \vspace{-2.25em}
    \caption{\label{fig:residual-init-ablation}%
        Residual initializations (green) drastically stabilize training of the LogicTreeNet compared to Gaussian initialization (orange).
    }
\end{wrapfigure}
Next, we consider the omission of \textit{or} pooling. We can observe that the accuracy drops by $3.5\%$ when removing \textit{or} pooling, demonstrating its importance.
Setting weight decay to $0$ causes a small reduction in accuracy by $1\%$.
Allowing each tree to use 8 channels as the input, rather than just 2, reduces the accuracy ($1.4\%$) because it is better to enforce the ability to perform comparisons within one channel at different $x,y$ locations in the kernel. 
However, the more important effect of using only 2 input channels is the resulting improved routing in hardware design layouts.

Finally, we ablate the proposed residual initializations. 
We can observe in the table that the accuracy drops by almost $9\%$ without residual initializations.
This means the that the Gaussian initialization are almost unusable for such deep networks.
In Figure~\ref{fig:residual-init-ablation}, we display the test accuracy during training and observe that, without our residual initializations, training does not converge and is quite unstable.

\begin{figure}[h!]
    \centering
    \vspace{.25em}
    \includegraphics[trim={4mm 0mm 4mm 3mm},clip,width=0.495\linewidth]{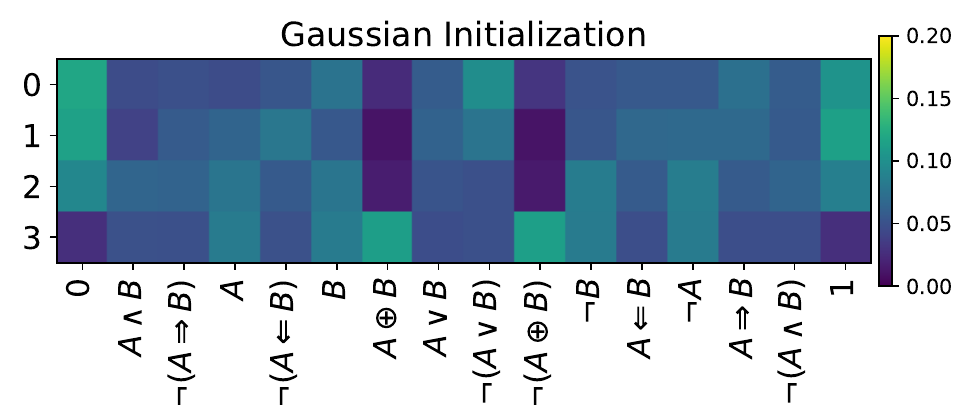}\hfill
    \includegraphics[trim={4mm 0mm 4mm 3mm},clip,width=0.495\linewidth]{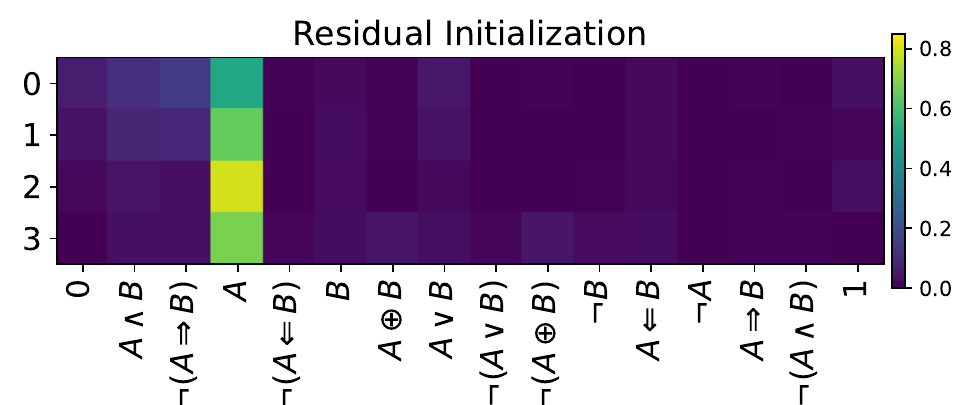}
    \caption{Distributions of choices of logic gates in a trained MNIST model, comparing Gaussian (left) and residual (right) initializations. The row number indicates the layer and the column indicates the logic gate.}
    \label{fig:dist-gates}
\end{figure}

We further ablate the effect of residual initialization on the distribution of gates in Figure~\ref{fig:dist-gates}.
Here, we can observe that residual initializations not only stabilize training, but also lead to the favorable inductive bias of many gates being the `$A$`, which is automatically reduced during logic simplification.

\vspace{-.5em}
\section{Conclusion}
\vspace{-.4em}

In this paper, we introduced convolutional differentiable logic gate networks with logic gate tree kernels, integrating a range of concepts from machine vision into differentiable logic gate networks.
In particular, we introduced residual initializations, which not only reduces loss of information in deeper networks, but also prevents vanishing gradients, enabling training of deeper LGNs than previously possible.
Further, we introduced logical \textit{or} pooling, which, combined with logic tree kernels, substantially improved training efficiency.
Our proposed CIFAR-10 architecture, LogicTreeNet, decreases model sizes {by factors of $\geq29\times$} compared to the SOTA while improving accuracy. 
Further, our inference stack demonstrates that convolutional LGNs can be efficiently executed on hardware. 
For example, on MNIST, our model improves accuracy while achieving {$160\times$} faster inference speed, and on CIFAR-10, our model improves inference speed by $1900\times$ over the state-of-the-art.
An interesting direction for future research is applying convolutional differentiable logic gate networks to computer vision tasks with continuous decisions like object localization.
We hope that our results motivate the community to adopt convolutional differentiable LGNs, especially for embedded and real-time applications where inference cost and speed matter most.

\vspace{-.3em}
\begin{ack}
\vspace{-.3em}
    This work was supported in part by the Federal Agency for Disruptive Innovation SPRIN-D. 
    CB~is~supported by the Land Salzburg within the WISS 2025 project IDA-Lab (20102-F1901166-KZP and 20204-WISS/225/197-2019). 
    SE is supported by the ARO~(W911NF-21-1-0125), the ONR~(N00014-23-1-2159), and the CZ~Biohub.
    We thank the reviewers for their supportive and helpful comments.
\end{ack}

\clearpage

\clearpage
\printbibliography 
\clearpage
\appendix

\section{Implementation Details}

\subsection{Model Architecture Details}
\label{apx:model}

In this section, we discuss the convolutional LGN architectures in detail.

\subsubsection{CIFAR-10 Architecture}

In the following, we describe the model for CIFAR-10 from Figure~\ref{fig:arch} layer by layer:

\begin{itemize}
\itemindent=-8pt
    \item A convolutional block with \texttt{k} kernels with a receptive field of size $3\times3$ and tree depth $d=3$, i.e., each kernel is a logic gate tree with seven logic gates, mapping 8 inputs to one output. 
    \item An \texttt{or} pooling layer with kernel size $2 \times 2$ and stride~2. ~[shape after layer: $\texttt{k}\times16\times16$]
    \item A convolutional block with \texttt{4*k} kernels with a receptive field of size $3\times3$ and depth $d=3$.
    \item An \texttt{or} pooling layer with kernel size $2 \times 2$ and stride~2. ~[shape after layer: $\texttt{4*k}\times8\times8$]
    \item A convolutional block with \texttt{16*k} kernels with a receptive field of size $3\times3$ and depth $d=3$.
    \item An \texttt{or} pooling layer with kernel size $2 \times 2$ and stride~2. ~[shape after layer: $\texttt{16*k}\times4\times4$]
    \item A convolutional block with \texttt{32*k} kernels with a receptive field of size $3\times3$ and depth $d=3$.
    \item An \texttt{or} pooling layer with kernel size $2 \times 2$ and stride~2. ~[shape after layer: $\texttt{32*k}\times2\times2$]
    \item Flattening the hidden state. ~[shape after flattening: $\texttt{128*k}$]
    \item Regular differentiable logic layer $\texttt{128*k}\to\texttt{1280*k}\,^{(*)}$.
    \item Regular differentiable logic layer $\texttt{1280*k}\to\texttt{640*k}\,^{(*)}$.
    \item Regular differentiable logic layer $\texttt{640*k}\to\texttt{320*k}\,^{(*)}$.
    \item GroupSum with 10 classes $\texttt{320*k}\to\texttt{10}$.
\end{itemize}
(All convolutional blocks are zero-padded with padding of size 1 to maintain respective shapes.)

$^{(*)}$: For~the B \& L size CIFAR models, we use $2\times$ as many gates in the final layers.

An additional implementation detail is that we can implement the last 3 layers in a fused fashion, using a single convolutional block/layer with depth $d=3$, and only a single kernel application, which is functionally equivalent and faster due to our fused kernels that are available for convolution.

\subsubsection{MNIST Architecture}
\label{apx:model-mnist}
The architecture for MNIST--like data sets has to account for the smaller input sizes ($28\times28$).
Thus, we use only 3 instead of 4 convolutional blocks for this architecture.

\begin{itemize}
\itemindent=-8pt
    \item A convolutional block with \texttt{k} kernels with a receptive field of size $5\times5$ and tree depth $d=3$, without padding.
    \item An \texttt{or} pooling layer with kernel size $2 \times 2$ and stride~2. ~[shape after layer: $\texttt{k}\times12\times12$]
    \item A convolutional block with \texttt{3*k} kernels with a receptive field of size $3\times3$ and depth $d=3$.
    \item An \texttt{or} pooling layer with kernel size $2 \times 2$ and stride~2. ~[shape after layer: $\texttt{3*k}\times6\times6$]
    \item A convolutional block with \texttt{9*k} kernels with a receptive field of size $3\times3$ and depth $d=3$.
    \item An \texttt{or} pooling layer with kernel size $2 \times 2$ and stride~2. ~[shape after layer: $\texttt{9*k}\times3\times3$]
    \item Flattening the hidden state. ~[shape after flattening: $\texttt{81*k}$]
    \item Regular differentiable logic layer $\texttt{81*k}\to\texttt{1280*k}\,^{(*)}$.
    \item Regular differentiable logic layer $\texttt{1280*k}\to\texttt{640*k}\,^{(*)}$.
    \item Regular differentiable logic layer $\texttt{640*k}\to\texttt{320*k}\,^{(*)}$.
    \item GroupSum with 10 classes $\texttt{320*k}\to\texttt{10}$.
\end{itemize}

$^{(*)}$: For~the S \& M size MNIST models, we use $2\times$ as many gates in the final layers.

\subsection{Training Details}
\label{apx:training}

In Table~\ref{tab:hyperparameters}, we summarize the hyperparameters for each model architecture configuration.
We observe that the hyperparameter that depends the most on the data set is the learning rate~$\eta$. 
The temperature~$\tau$ and thus the range of attainable outputs $n_{\ell\ell/c}/\tau$ has a minor dependence on the data set.
We use weight decay only for the CIFAR-10 models as it does not yield advantages for the smaller MNIST models.
We note that convergence, when training with weight decay, is generally slightly slower but leads to slightly better models. 
Models trained wiht weight decay tend to have more gates.

\begin{table*}[h!]
    \centering
    \captionof{table}{
        Hyperparameters for each model and data set:
        softmax temperatures~$\tau$, learning rates~$\eta$, weight decays~$\beta$, and batch sizes~$\mathit{bs}$.
        For reference to show the relationship to~$\tau$, we include the number of output neurons in the last layer per class $n_{\ell\ell/c}$. The range of attainable class scores is $[0, n_{\ell\ell/c}/\tau]$. 
    \label{tab:hyperparameters}
    }
    {\small
    \begin{tabular}{lc|rrrrr|rrr}
    \toprule
        Data set                      &              & \multicolumn{5}{c|}{CIFAR-10}                   & \multicolumn{3}{c}{MNIST}      \\[.35em]
        Model identifier              &              &  S & M & B & L & G                              & S & M & L  \\
        Model scale&\texttt{k}                       & 32     & 256    & 512 & 1\,024 & 2\,560         & 16 & 64 & 1\,024  \\
    \midrule
        Temperature&$\tau$                           & 20     & 40     & 280    & 340    & 450         & 6.5 & 28 & 35                           \\
        Learning rate&$\eta$                         & 0.02   & 0.02   & 0.02   & 0.02   & 0.02        & 0.01 & 0.01 & 0.01                  \\
        Weight decay&$\beta$                         & 0.002  & 0.002  & 0.002  & 0.002  & 0.001       & 0 & 0 & 0                               \\
        Batch size&$\texttt{bs}$                     & 128    & 128    & 128    & 128    & 128         & 512 & 256 & 128                      \\
        Output gate factor&$\texttt{ox}$             & 1      & 1      & 2      & 2      & 1           & 2   & 2   & 1                        \\
        \# input bits &                              & 2      & 2      & 5      & 5      & 5           & 1   & 1   & 1                        \\
    \midrule
        Outputs/class &\kern-.5em$n_{\ell\ell/c}$    & 1K     & 8K    & 32K     & 64K    & 80K         & 1K & 4K & 8K                  \\
        Max score &\kern-1.4em$n_{\ell\ell/c}/\tau$  & 51     & 205   & 117     & 193    & 182         & 158 & 146 & 234                  \\
    \bottomrule
    \end{tabular}
    }
\end{table*}

For the loss, we use different softmax temperatures $\tau$ depending on the model size.
We observe two important relationships for choosing $\tau$: 
(i) $\tau$ depends on the number of output neurons (larger number of output neurons $\Rightarrow$ larger $\tau$) and 
(ii) $\tau$ depends on how certain the model will be after training on the respective dataset, i.e., for a hard task with low accuracy, we should choose a larger $\tau$, while, for an easier task with a higher accuracy, we should choose a smaller $\tau$.
The reason for this is that cross-entropy requires smaller logit variances if the model is less certain and requires larger logit variances if a prediction is certain.
A good rule of thumb during scaling is that the optimal temperature is proportional to the square-root of the number of output gates ($\tau^\star\propto \sqrt{n_{\ell\ell/c}}$).

For the CIFAR-10 B, L, and G models, we use a neural network teacher, supervising the class scores.
When using a teacher on the class score level, a good rule of thumb is to increase the softmax temperature by a factor of $\sqrt{2}$.

For CIFAR-10, we split the training data into 45\,000 training images and 5\,000 validation images, and evaluate every 2\,000 steps to select the best model.
For MNIST, we split the training data into 50\,000 training images and 10\,000 validation images, and evaluate every 5\,000 steps to select the best model.

\subsubsection{Memory Access Advantages through Fused Trees and Pooling}

Using logic gate trees as convolutional kernels with pooling allows for a substantial speedup during training.
For this, we fuse the entire tree as well as pooling into a single CUDA kernel operation.
The reason behind this is two-fold, illustrated for the case of depth $d=3$ and $2\times2$ pooling:
(i) after reading the necessary $32$ inputs, we perform $2\times2=4$ applications of the $7$ learnable logic gates comprising the tree.
Then, we apply the maximum t-conorm to pool the $4$ outputs of the $4$ tree applications to a single output value.
Here, we do not need to read the intermediate results from memory but can instead keep them in registers.
This prevents $28$ additional memory read operations.
(ii) as each set of $4$ tree applications only has a single output after pooling, it is sufficient to write only this individual output (as well as the index of the pooling operation) to memory, saving $28$ memory write operations, which are expensive.
Further, this also reduces the memory footprint of training by a factor of around $10\times$.
This procedure requires recomputing selected intermediate values within each block during the backward pass; however, the memory access savings offset this small additional computational cost.

\subsubsection{Computational Requirements}

The typical training time per epoch for the L model on a single NVIDIA RTX 4090 GPU is 30~seconds.
Noteworthy is that $d=3$ kernels are typically bottlenecked by CUDA compute cores, whereas $d=2$ and $d=1$ kernels are bottlenecked by memory bandwidth.
While we explored $d=4$ kernels, they (when fused) are very expensive ($>10\times$) due to register pressure. 
Generally, but with very limited exploration, simply going from $d=3$ to $d=4$ did not improve performance / gate.
$d=4$ kernels can also be expressed, without fusing, using 2 logic gate tree layers; however, with this the memory consumption during training increases ($\approx10\times$) which becomes a bottleneck.

\subsection{Inference Details}

\begin{wrapfigure}[18]{r}{0.33\textwidth}
    \centering
    \vspace{-1.25em}
    \includegraphics[width=.875\linewidth]{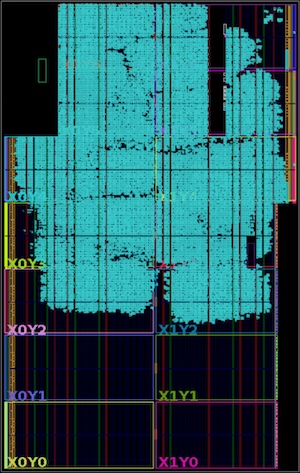}
    \caption{
        MNIST model (M).
    \label{fig:mnist-m-placement}
    }
\end{wrapfigure}
For efficient hardware implementations, routing is an important consideration and the ability to place the LGN without congestion is paramount.
Due to the sparsity of differentiable logic gate networks, limiting the choice of connections to a reasonable degree does not negatively affect accuracy.
Accordingly, we select connections such that the model could be split into $\texttt{k}/8$ separated models that are only recombined at the stage of output gates after accumulation, akin to using grouped convolutions with a constant number of groups throughout the network.
This prevents congestions without reducing the accuracy. 
Additional routing restrictions can straightforwardly be implemented in differentiable LGNs without incurring performance penalties.

We illustrate the placement of our MNIST model (M) on a Xilinx XC7Z045 FPGA in Figure~\ref{fig:mnist-m-placement}.

We developed scalable logic synthesis tools that simplify the logic gate networks after training.
For example, for our large MNIST model (L), during training, the main network has $3\,216\,128$ gates.
After training, in the discrete LGN, many of these gates are trivial (e.g., `$A$` or constant) or not connected, and also further simplifications are possible.
After logic synthesis, the number of logic gates was $697\,758$. Herein, the full convolutional nature of the network is maintained (and gates that have zero padded input, are still counted as full gates.)
For the group sum operation, we use a tree adder that asymptotically uses 7 gates per output.

\subsection{Train / Test Accuracy and Discretization Error}
\label{apx:train-test-discrete}

\begin{wrapfigure}[19]{r}{0.56\textwidth}
    \centering
    \kern-1em\includegraphics[width=1.04\linewidth]{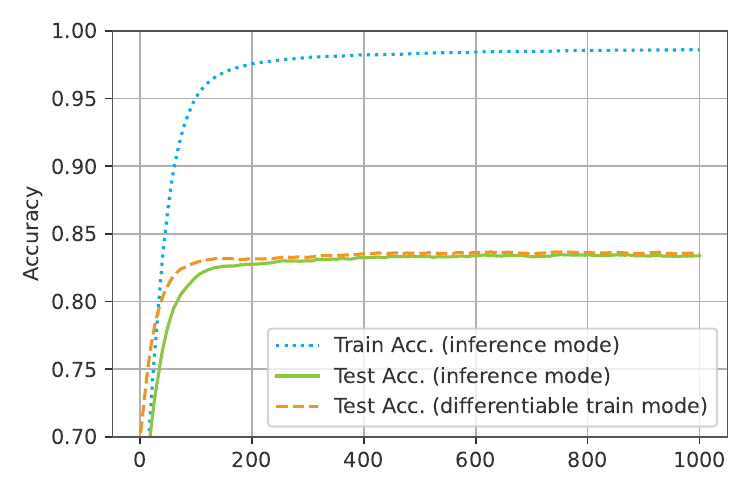}
    \vspace{-1.5em}
    \caption{
        CIFAR-10 training and test accuracy plot. 
        The discretization error, i.e., the difference between the inference (hard) mode accuracy and the differentiable training mode accuracy is very small during late training.
    \label{fig:train-test-cifar-10}
    }
\end{wrapfigure}

In Figure~\ref{fig:train-test-cifar-10}, we plot the training and test accuracies for training a convolutional LGN on CIFAR-10~\cite{krizhevsky2009cifar10}.
Here, we can observe that the discretization error, i.e., the difference between the inference (hard) mode accuracy and the differentiable training mode accuracy is very small during late training.
During early training, the discretization error is more substantial because the method first learns a ``smooth'' differentiable LGN (with high levels of uncertainty in the individual logic gate choices), which later converges to discrete choices for each logic gate.
The discretization step chooses the logic gate with the highest probability for inference mode.
Accordingly, during early training, the discretization causes larger changes, negatively affecting accuracy, while during late training, the discretization barely causes any changes and therefore does not considerably affect accuracy. 
We note that there is no noticeable overfitting behavior.

\subsection{Ablation of the Residual Initialization Hyperparameter \protect$z_3$}

\begin{wrapfigure}[11]{r}{0.47\textwidth}
    \centering
    \vspace{-1.6em}
    \includegraphics[width=\linewidth]{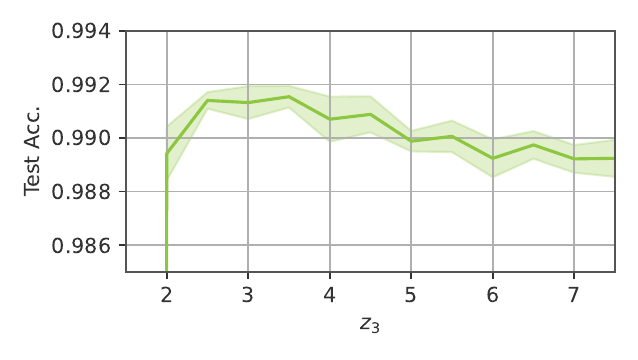}
    \vspace{-2em}
    \caption{
    Test accuracy of an MNIST model with different choices of $z_3$ for the residual initialization, in steps of 0.5. 
Averaged over 5 seeds.
    }
    \label{fig:z3-ablation}
\end{wrapfigure}
In Figure~\ref{fig:z3-ablation}, we ablate the choice of $z_3$, which is the hyperparameter that indicates how strong the residual initialization is applied. 
We illustrate the ablation for an MNIST model.
The model performs well, when $z_3\geq 2$ ($z_3 = 1.5$ is included but reaches only 13\%.) 
While $z_3 = 5$ is not the optimal choice for this particular model and training length, we have observed
that, for larger models as well as for longer trainings, larger $z_3$ tend to be favorable.
For example, on CIFAR-10, with the greater model depth, a $z_3$ of $2$ is too small and prevents training, so we generally recommend using $z_3=5$.

\section{Additional BMAC and BNN Discussions}

\subsection{BMACs in Logic}

When translating a BNN into an LGN for inference on FPGAs, BNNs require BMACs (multiply-accumulate), which are often expressed via an XNOR ($\neg(a\oplus b)$) for the multiplication and a bitcount operation for the accumulation.
In the case of $n$ input bits, the necessary $n$ MAC can be expressed using $\mathcal{O}(n)\approx 8\times n$ logic gates: $n$ logic gates are necessary for the XNORs and $\approx 7n$ logic gates for the accumulating bitcount.
Further, this process adds a delay of $\mathcal{O}(\log n)$ to the logic circuit.
This means that a BMAC is not one logical operation but instead typically requires around $8$ binary logical operations (not accounting for additional thresholding or potentially batch-normalization operations).

\subsection{Floats in BNNs}

BNNs typically use 1-bit input activation and 1-bit~weight quantization, but no output activation quantization~\cite{santos2023redbit}.
This means that only the multiplications within a matrix multiplication are binary, while the remainder of the respective architectures is typically non-binary~\cite{santos2023redbit}. 
As in some residual BNN approaches~\cite{tu2022adabin,gong2019differentiable,qin2020forward,qiu2022rbnn,zhou2016dorefa,shen2020balanced}, the residuals are not quantized, they require an additional FLOP (or integer operation) overhead between the layers; due to the cost of FLOPs in logic gates ($>$1000 binary OPs), effective deployment on a logic level has not been demonstrated for these unquantized residual approaches.
Quantizing the residuals typically comes at a substantial accuracy penalty as demonstrated, e.g., by Ghasemzadeh~\textit{et al.}~\cite{ghasemzadeh2018rebnet}.
Thus, as these networks use full-precision residuals, a fair comparison is not applicable.
Still, float-residual BNN approaches have important implications for speeding up GPU inference if (i) large matrix multiplications are necessary and (ii) the savings during the binary matrix multiplication itself outweigh the costs of converting between bit-wise representations and floats/integer representations; however, float-residual BNNs are not suitable for efficient logic gate based inference in hardware, e.g., on FPGAs or ASICs.

\section{List of Assets}

\begin{itemize}
    \item CIFAR-10~\cite{krizhevsky2009cifar10} ~~[different open licenses]
    \item MNIST~\cite{lecun2010mnist} ~~[CC License]
    \item PyTorch~\cite{2019-PyTorch} ~~[BSD 3-Clause License]
\end{itemize}

\end{document}